\documentclass[runningheads,a4paper]{llncs}
\usepackage[T1]{fontenc}
\usepackage{newtxtext}
\usepackage[varvw]{newtxmath}
\usepackage{hyperref}
\usepackage{url}
\usepackage{booktabs}
\usepackage{microtype}
\usepackage{xcolor}
\usepackage{tikz}
\usetikzlibrary{arrows.meta,positioning,shapes.multipart,fit,calc}
\usepackage{tabularx}
\usepackage{graphicx}
\newcommand{\AnnAccN}{39}
\newcommand{\AnnAwayDiff}{$-$18.6 [$-$31.0, $-$6.8]}
\newcommand{\AnnFactsHumApp}{$-$24.6 [$-$40.4, $-$9.6]}
\newcommand{\AnnFactsHumDeny}{$-$25.0 [$-$43.3, $-$8.3]}
\newcommand{\AnnPrecAgree}{70}
\newcommand{\AnnPrecedents}{70}
\newcommand{\AnnTargetNotClear}{35}
\newcommand{\AnnTargets}{40}
\newcommand{\AnnTowardDiff}{+20.5 [10.0, 30.3]}
\newcommand{\AxApproveOmittedSeeds}{0/6}
\newcommand{\AxDenyApproveSeeds}{6/6}
\newcommand{\AxDenyDenySeeds}{0/6}
\newcommand{\AxDenyOmittedSeeds}{6/6}
\newcommand{\AxNoneSeeds}{3/6}
\newcommand{\BordCalls}{34}
\newcommand{\BordIn}{24}
\newcommand{\BordLIn}{+35.1 [23.8, 46.4]}
\newcommand{\BordLOut}{+41.5 [28.9, 51.9]}
\newcommand{\BordOut}{41}
\newcommand{\BtwOpusHaikuLPt}{$-$33.7}
\newcommand{\BtwSolLunaSixLPt}{$-$35.4}
\newcommand{\CThreeDeep}{+53.7 [40.7, 66.7]}
\newcommand{\CThreeDeepPt}{+53.7}
\newcommand{\CThreeHaiku}{+30.9 [16.9, 45.7]}
\newcommand{\CThreeN}{34}
\newcommand{\DownOpusSr}{20}
\newcommand{\ETwoLabelAsstEst}{46.3}
\newcommand{\ETwoLabelToolEst}{40.8}
\newcommand{\ILBase}{+39.0 [28.0, 48.7]}
\newcommand{\ILIgnore}{+29.1 [16.8, 41.9]}
\newcommand{\ILWarn}{+21.8 [9.2, 34.6]}
\newcommand{\IdLIgnore}{$-$9.9 [$-$17.1, $-$3.2]}
\newcommand{\IdLWarn}{$-$17.2 [$-$25.4, $-$9.4]}
\newcommand{\ModDecisions}{2,730}
\newcommand{\ModFCells}{12}
\newcommand{\ModFMax}{41}
\newcommand{\ModFMin}{13}
\newcommand{\ModSeeds}{3}
\newcommand{\ModSolHumNone}{94.9}
\newcommand{\NCandidates}{310}
\newcommand{\NTargetsPublic}{65}
\newcommand{\ODDF}{$-$13.6 [$-$26.7, 0.0]}
\newcommand{\ODDFHolm}{0.0096}
\newcommand{\ODDFPolP}{0.0234}
\newcommand{\ODDL}{+33.3 [23.6, 42.6]}
\newcommand{\ODDLHolm}{0.0009}
\newcommand{\ODDLPolP}{0.0039}
\newcommand{\ODFDonor}{$-$16.7 [$-$29.7, $-$2.7]}
\newcommand{\ODFDonorHolm}{0.0004}
\newcommand{\ODFDonorPolP}{0.0273}
\newcommand{\ODFOwn}{$-$30.3 [$-$43.7, $-$17.4]}
\newcommand{\ODFOwnHolm}{0.0009}
\newcommand{\ODFOwnPolP}{0.0039}
\newcommand{\ODLDonor}{+6.3 [$-$1.4, 14.0]}
\newcommand{\ODLDonorHolm}{0.0316}
\newcommand{\ODLDonorPolP}{0.1250}
\newcommand{\ODLOwn}{+39.6 [29.8, 49.0]}
\newcommand{\ODLOwnHolm}{0.0009}
\newcommand{\ODLOwnPolP}{0.0039}
\newcommand{\ODSeeds}{6}
\newcommand{\RFCellN}{390}
\newcommand{\RFFacts}{$-$23.8 [$-$37.0, $-$12.5]}
\newcommand{\RFFactsPolP}{0.0039}
\newcommand{\RFFactsPt}{$-$23.8}
\newcommand{\RFLabel}{+39.1 [28.7, 49.0]}
\newcommand{\RFLabelEst}{39.1}
\newcommand{\RFLabelPolP}{0.0039}
\newcommand{\RFLabelPt}{+39.1}
\newcommand{\RFRateApproveStrong}{40.0}
\newcommand{\RFRateApproveWeak}{76.7}
\newcommand{\RFRateNone}{51.5}
\newcommand{\RFRateOmittedStrong}{34.6}
\newcommand{\RFRateOmittedWeak}{58.5}
\newcommand{\RFRateRejectStrong}{6.4}
\newcommand{\RFRateRejectWeak}{32.1}
\newcommand{\RFSeeds}{6}
\newcommand{\RetDSLFullPt}{+71.0}
\newcommand{\RetDSLLooPt}{+25.1}
\newcommand{\RetDSSeeds}{6}
\newcommand{\RetDSWFullEst}{24.9}
\newcommand{\RetDSWFullPt}{+24.9}
\newcommand{\RetDSWLooEst}{15.6}
\newcommand{\RetDSWLooPt}{+15.6}
\newcommand{\RetHumDSFlipMinusRef}{$-$23.9 [$-$41.5, $-$6.4]}
\newcommand{\RetHumOpusFlipMinusRef}{+12.0 [4.3, 21.4]}
\newcommand{\RetHumOpusRefMinusNone}{$-$19.7 [$-$30.8, $-$9.4]}
\newcommand{\RetHumSolFlipMinusRef}{+1.7 [$-$5.1, 8.5]}
\newcommand{\RetHumSolRefMinusNone}{$-$15.4 [$-$26.5, $-$6.0]}
\newcommand{\RetOwnAll}{61}
\newcommand{\RetStore}{780}
\newcommand{\SRFacts}{$-$12.1 [$-$27.1, 2.9]}
\newcommand{\SRLabel}{+51.9 [40.6, 60.9]}
\newcommand{\SRLabelPt}{+51.9}
\newcommand{\TargetsPerPolicyMax}{11}
\newcommand{\TargetsPerPolicyMin}{5}
\newcommand{\ThinkOff}{+55.1 [43.2, 66.2]}
\newcommand{\ThinkOn}{+37.5 [25.8, 50.0]}
\newcommand{\UpOpusSr}{11}
\newcommand{\XAOrigEmpty}{$-$23.4 [$-$32.7, $-$14.1]}
\newcommand{\XASrEmpty}{$-$17.1 [$-$35.0, $-$1.6]}

\hypersetup{hidelinks}
\begin{document}
\title{How Retrieved Past Cases Influence LLM Approval Decisions: Recorded Outcomes and Case Facts}
\titlerunning{How Retrieved Past Cases Influence LLM Approval Decisions}
\author{Junwei Yao\inst{1} \and Yanxian Chen\inst{1} \and Zike Yang\inst{1} \and Yongxi Zhou\inst{2} \and Jianan Liu\inst{2} \and Jiaxi Wen\inst{1} \and Zhijun Wang\inst{1}}
\authorrunning{J. Yao et al.}
\institute{Independent Researcher, Sunnyvale, CA, United States\\ \email{yao.junw@northeastern.edu} \and Independent Researcher, Austin, TX, United States}
\maketitle
\begin{abstract}
Decision-support systems built on language models often show the model similar past cases next to the written
policy. A past case has two parts: its facts and the decision recorded for it (its label). We propose an audit that
measures which part moves the model's decision. The test requests are refund, warranty and claim requests under nine
paraphrased public policies, selected as close calls (requests that DeepSeek V4.1 Flash decides both ways). Each is shown with three past cases
that are near-copies of it, and we change one part at a time. Changing only the labels from DENY to APPROVE raises
DeepSeek's approval rate by \RFLabelEst{} percentage points. An instruction not to rely on recorded decisions shrinks
this label effect without removing it, and when a keyword search picks the cases from a store, flipped stored labels move DeepSeek's
decisions toward them. Three Claude and three GPT models decided the same requests. Claude Haiku 5.5, Claude Sonnet 5.5 and GPT-6 Luna also
follow the labels; for the largest, Claude Opus 5.5 and GPT-6.1 Sol, no positive label effect was detected, and Sol moved away from the labels with one set of variants. The
facts push the other way. With the labels removed, past cases whose facts call for approval lead to fewer approvals
than cases whose facts call for denial, by \ModFMin{} to \ModFMax{} points in all six Claude and GPT models. Cases shown without a
decision should not be assumed neutral for any of the models tested.
\keywords{Case-based reasoning \and Retrieval-augmented generation \and Large language models \and Decision support \and Trustworthy AI}
\end{abstract}

\section{Introduction}
\label{sec:intro}

A refund desk, a warranty service or a claims team has a policy, a stream of requests and a record of past decisions.
Case-based reasoning retrieves similar solved cases and reuses their outcomes for a new problem~\cite{aamodt1994cbr}.
When a language model decides the requests, the same idea becomes a retrieval step that places similar past decisions
in the model's context~\cite{wiratunga2024cbrrag,chen2025caselaw} next to the policy text. A stored case
carries two kinds of knowledge, its facts and the outcome recorded for it, and the recorded outcome can be wrong. We ask whether the recorded outcomes and the facts of past cases act on the model's
decision in the same way. For the designer of such a system the answer decides whether to check the stored
outcomes, the retrieved cases themselves, or both.

We call the request under decision the \emph{target} and the past cases shown with it its \emph{precedents}. Each
precedent is a fact sheet followed by its recorded decision, its \emph{label}. Figure~\ref{fig:example} shows the
setting. A guest asks Airbnb for a refund because the hot water in a five-night rental runs out after about ten minutes
of showering. The paraphrased policy covers an advertised amenity that does not function but excludes minor
inconveniences, so the request is a close call. Without precedents, DeepSeek V4.1 Flash approved it in
\AxNoneSeeds{} decisions. Its precedents are \emph{close variants}, near-copies of the request with a few fields
changed. Next to three deny-side variants, DeepSeek approved the request in \AxDenyDenySeeds{} decisions when they were
labeled DENY and in \AxDenyApproveSeeds{} when the same cases were labeled APPROVE.

\begin{figure}[t]
\centering
\footnotesize
\begin{tabularx}{\linewidth}{@{}X@{}}
\toprule
\textbf{Policy (Airbnb Rebooking and Refund Policy, paraphrased; abridged).} 3. A special amenity advertised
in the listing that is not present or does not function is covered; minor inconveniences are not. 4. The guest must
report the issue within 72 hours of discovering it. 5. Issues caused by the guest are not covered. \dots\\[2pt]
\textbf{Target request.} Five nights, apartment for four; hot water runs out after about 10 minutes of showering;
found and reported on day 1 with a video; the host says the tank is small and refills in an hour; the guest stayed.\\[2pt]
\textbf{Deny-side close variants.} The same stay, except: (1) the shower head sprays slightly to one side; (2) the
hot water runs out because the guests left the hot tap running for an hour; (3) the issue is reported 3 weeks after
checkout, with no evidence.\\[2pt]
\textbf{Approve-side close variants.} The same stay, except: (1) the host canceled the day before check-in; (2) no
hot water at all for the whole stay in winter, and the guest left on day 2; (3) no running water at all, and the
guest left the same day.\\
\midrule
\end{tabularx}
\begin{tabularx}{\linewidth}{@{}>{\raggedright\arraybackslash}p{3.9cm}c>{\raggedright\arraybackslash}X@{}}
Precedents shown & Approved & Sampled reason (verbatim, abridged)\\
\midrule
None & \AxNoneSeeds{} & \\
Deny-side, DENY labels & \AxDenyDenySeeds{} & \\
Deny-side, APPROVE labels & \AxDenyApproveSeeds{} & ``The advertised hot water is a special amenity that does not function (criterion 3) \dots''\\
Deny-side, no labels & \AxDenyOmittedSeeds{} & ``A listed amenity (hot water for normal showering) not functioning is a significant difference under criterion 3 \dots''\\
Approve-side, no labels & \AxApproveOmittedSeeds{} & ``\dots\ is a minor inconvenience rather than a missing or non-functioning advertised amenity \dots''\\
\bottomrule
\end{tabularx}
\caption{Relabeling three precedents raises approval of a close-call request; swapping in approve-side facts lowers
it. DeepSeek V4.1 Flash. ``Approved'' counts approvals over six repeated decisions per condition; reasons are shown for three conditions, each from one of its six decisions.}
\label{fig:example}
\end{figure}

\begin{figure}[t]
\centering
\resizebox{\linewidth}{!}{\resizebox{\linewidth}{!}{%
\begin{tikzpicture}[
  font=\footnotesize,
  box/.style={draw, rounded corners=1pt, align=center, inner sep=3pt},
  pc/.style={rectangle split, rectangle split parts=2, draw, rounded corners=1pt, align=left, text width=3.0cm, inner sep=3pt},
  title/.style={font=\footnotesize\bfseries, anchor=west},
  arr/.style={-{Latex[length=2mm]}, thick}
]
\node[title] at (5.2,3.9) {(a) DeepSeek's approval rate (\%)};
\begin{scope}[shift={(7.1,0)}, x=0.06cm,
  mk/.style={inner sep=0pt, minimum size=5pt, draw, thick},
  val/.style={font=\scriptsize, below=3pt, fill=white, inner sep=1pt}]
\draw[gray] (0,0.7) -- (100,0.7);
\foreach \t in {0,25,50,75,100} {\draw[gray] (\t,0.7) -- (\t,0.62) node[below, font=\scriptsize, text=black] {\t};}
\draw[dashed, gray] (\RFRateNone,0.7) -- (\RFRateNone,3.4) node[below right, font=\scriptsize, text=black] {no precedents (\RFRateNone\%)};
\node[anchor=east, align=right] at (-5,2.9) {Approve-side\\facts};
\node[anchor=east, align=right] at (-5,1.4) {Deny-side\\facts};
\draw[gray] (\RFRateRejectStrong,2.9) -- (\RFRateApproveStrong,2.9);
\draw[gray] (\RFRateRejectWeak,1.4) -- (\RFRateApproveWeak,1.4);
\node[val] at (\RFRateRejectStrong,2.9) {\RFRateRejectStrong}; \node[mk, fill=black] at (\RFRateRejectStrong,2.9) {};
\node[val, anchor=north east, xshift=4pt] at (\RFRateOmittedStrong,2.9) {\RFRateOmittedStrong}; \node[mk, circle, fill=white] (cs) at (\RFRateOmittedStrong,2.9) {};
\node[font=\scriptsize, right=3pt, fill=white, inner sep=1pt] at (\RFRateApproveStrong,2.9) {\RFRateApproveStrong}; \node[mk, fill=white] at (\RFRateApproveStrong,2.9) {};
\node[val] at (\RFRateRejectWeak,1.4) {\RFRateRejectWeak}; \node[mk, fill=black] at (\RFRateRejectWeak,1.4) {};
\node[val] at (\RFRateOmittedWeak,1.4) {\RFRateOmittedWeak}; \node[mk, circle, fill=white] (cw) at (\RFRateOmittedWeak,1.4) {};
\node[val] at (\RFRateApproveWeak,1.4) {\RFRateApproveWeak}; \node[mk, fill=white] at (\RFRateApproveWeak,1.4) {};
\draw[arr, thin] (\RFRateRejectStrong,3.15) -- node[above, font=\scriptsize, inner sep=1pt] {$L$} (\RFRateApproveStrong,3.15);
\draw[arr, thin] (\RFRateRejectWeak,1.65) -- node[above, font=\scriptsize, inner sep=1pt] {$L$} (\RFRateApproveWeak,1.65);
\draw[densely dotted] (\RFRateOmittedStrong,2.35) -- (\RFRateOmittedStrong,2.15);
\draw[densely dotted] (\RFRateOmittedWeak,1.5) -- (\RFRateOmittedWeak,2.15);
\draw[arr, thin] (\RFRateOmittedWeak,2.15) -- node[above, font=\scriptsize, inner sep=1pt] {$F$} (\RFRateOmittedStrong,2.15);
\node[mk, fill=black] at (-18,-0.05) {}; \node[anchor=west, font=\scriptsize] at (-16,-0.05) {DENY labels};
\node[mk, circle, fill=white] at (22,-0.05) {}; \node[anchor=west, font=\scriptsize] at (24,-0.05) {no labels};
\node[mk, fill=white] at (58,-0.05) {}; \node[anchor=west, font=\scriptsize] at (60,-0.05) {APPROVE labels};
\end{scope}
\node[title] at (13.9,3.9) {(b) Whose precedents are shown};
\begin{scope}[card/.style={draw, rounded corners=1pt, minimum width=0.62cm, minimum height=0.5cm, inner sep=1pt, font=\footnotesize},
  own/.style={card, fill=black!18}, other/.style={card, fill=white}]
\node[anchor=west, font=\footnotesize\bfseries] at (14.0,3.3) {Own};
\node[own] (o1) at (14.35,2.6) {$A'$}; \node[own] (o2) at (15.05,2.6) {$A''$}; \node[own] (o3) at (15.75,2.6) {$A'''$};
\node[own, minimum width=2.02cm] (ot) at (15.05,1.35) {request $A$};
\draw[arr] (o2.south) -- (ot.north);
\node[align=center, font=\scriptsize] at (15.05,0.55) {near-copies of\\the request itself};
\node[anchor=west, font=\footnotesize\bfseries] at (16.9,3.3) {Donor};
\node[other] (d1) at (17.25,2.6) {$B'$}; \node[other] (d2) at (17.95,2.6) {$B''$}; \node[other] (d3) at (18.65,2.6) {$B'''$};
\node[own, minimum width=2.02cm] (dt) at (17.95,1.35) {request $A$};
\draw[arr] (d2.south) -- (dt.north);
\node[align=center, font=\scriptsize] at (17.95,0.55) {near-copies of another\\request $B$, same policy};
\end{scope}
\end{tikzpicture}}}
\caption{Measures and precedent sources. (a) Approval by DeepSeek
V4.1 Flash on the public policies, original variants (Table~\ref{tab:main}): each row is one set of facts, each marker one way of labeling the three cases. Each arrow marked $L$ is the DENY-to-APPROVE shift within one row, and $L$ is the average of the two; $F$ is the shift between the rows without labels.
(b) Own precedents $A'$--$A'''$ are close variants of the target $A$; donor precedents $B'$--$B'''$ are close variants
of another target $B$ of the same policy.}
\label{fig:design}
\end{figure}

We measure two effects (Fig.~\ref{fig:design}). The \emph{label effect} $L$ is the change in the target's approval rate, in percentage points,
when the same three precedents carry APPROVE instead of DENY labels. For the facts, each target has two groups of three
close variants: \emph{deny-side} variants, whose facts call for denial under the rules, and \emph{approve-side}
variants, whose facts call for approval. The \emph{facts contrast} $F$ is the change in approval when the approve-side
variants replace the deny-side ones, both shown without labels. In Figure~\ref{fig:example} it is negative. The
request was approved in \AxDenyOmittedSeeds{} decisions next to the unlabeled deny-side variants and in
\AxApproveOmittedSeeds{} next to the unlabeled approve-side ones. Next to variants with a canceled stay, no hot water
or no water at all, the model calls ten minutes of hot water a minor inconvenience.

The two contrasts form an audit that an operator can run with the system's own model, policy and case store. We run
it with DeepSeek V4.1 Flash on nine public policies and repeat the swaps on six Claude and GPT models and under six
fictional policies, to answer four research questions. \textbf{RQ1:} Do the recorded labels move the decision, in
which models, and does an instruction remove the effect? \textbf{RQ2:} Do the facts of unlabeled precedents move the
decision, and in which direction? \textbf{RQ3:} Does the label effect depend on the precedents being close variants of
the request? \textbf{RQ4:} Does the label effect appear when a retriever chooses the cases from a store, and do wrong
stored labels propagate?

Labels pull the decision toward themselves in four of the seven models (DeepSeek V4.1 Flash, Claude Haiku 5.5, Claude Sonnet 5.5, GPT-6 Luna), by \RFLabelEst{} points for
DeepSeek, and for DeepSeek an instruction lowers but does not remove the effect; for the largest Claude and GPT models no positive
label effect was detected (RQ1). Unlabeled approve-side precedents give lower approval than deny-side ones in all six
Claude and GPT models (RQ2). For DeepSeek, close variants of another request carry a much smaller label effect (RQ3). With a retriever,
flipping the stored labels moves DeepSeek's decisions \RetDSWFullEst{} points toward them, and \RetDSWLooEst{} points when the store holds no variants of the target (RQ4). Our contributions are
an audit that measures, with paired interventions under written rules, the response to precedent labels and the response to unlabeled case facts; evidence from seven models that the two responses differ between models, with models that do not follow the labels still moved by the facts; and checks for the design of such systems (Section~\ref{sec:implications}).

\section{Related work}
\label{sec:related}

\subsubsection{Case-based reasoning and retrieval for decisions.}
Case-based reasoning solves a new problem by retrieving similar past cases and reusing their solutions, revising them
where they do not fit~\cite{aamodt1994cbr}. Retrieval-augmented generation gives a language model retrieved text at
decision time~\cite{lewis2020rag}, and CBR-RAG combines the two, retrieving past cases for legal question answering~\cite{wiratunga2024cbrrag}. These designs reuse a retrieved case's outcome as a signal for the new
case. We test how a language model weighs a retrieved case's outcome against its facts when the written rules are the
authority, and what happens when stored outcomes are wrong. Recasens~et~al.~\cite{recasens2026bias} show that biases in the in-context examples a language model is given
propagate into the tabular data it generates, a related risk in another task. Case-base maintenance removes or edits stored cases that would mislead retrieval, for instance noisy cases
whose neighbors disagree with them~\cite{wilson1972asymptotic,brighton2002advances,delany2004analysis}, and deletion can also limit retrieval cost while preserving competence~\cite{smyth1995remembering}; we measure how far a
wrong stored outcome moves a language model's decision. Retrieval corpora can also be poisoned on purpose to steer
generation~\cite{li2026cparag}; the wrong outcomes we study need no attacker.

\subsubsection{Earlier judgments and anchoring.}
Temkit~\cite{temkit2026amel} shows that the polarity of earlier conversation turns shifts later LLM judgments of a fixed
item across 12 models, most on items the model is uncertain about. Chen~et~al.~\cite{chen2024threshold} find threshold priming in
LLM relevance assessment: documents judged after highly relevant documents receive lower scores. This contrast from
earlier items' content goes in the same direction as our facts contrast $F$. Echterhoff~et~al.~\cite{echterhoff2024cognitive} measure
sequential and other cognitive biases in decisions such as student admissions, and Kapetanovic~et~al.~\cite{kapetanovic2026anchoring}
show that a prior score for the same item anchors LLM judges. Wrappers that keep the content can flip LLM safety-judge verdicts~\cite{zhou2026style}, and scoring a single canonical prompt underestimates unsafe compliance across surface forms of a request~\cite{zhou2026canonical}. We vary the recorded outcome and the facts of the same retrieved cases independently.

\subsubsection{Labels in demonstrations.}
Wei~et~al.~\cite{wei2023larger} show that larger models can follow flipped in-context labels, while Min~et~al.~\cite{min2022rethinking}
find that random demonstration labels often matter little for classification accuracy; Yoo~et~al.~\cite{yoo2022ground} show that
how much correct labels matter depends on the setting. Demonstrations similar to the test input help~\cite{liu2022makes}, and Lyu~et~al.~\cite{lyu2023zicl} find that predictions copy the labels of demonstrations whose inputs
closely resemble the test input, also when those labels are wrong. Zhao~et~al.~\cite{zhao2021calibrate} and
Fei~et~al.~\cite{fei2023mitigating} describe and correct label biases from few-shot examples. In our setting retrieved records are shown next to written rules, and the model decides a single close-call target.

\subsubsection{Precedents as guidance.}
Chen and Zhang~\cite{chen2025caselaw} decide cases from retrieved past decisions and find that precedents align decisions with
ground truth better than high-level rules in most of the communities they study. Wu~et~al.~\cite{wu2023precedent} retrieve
precedents for legal judgment prediction. Both take the recorded outcomes of precedents as given. In factor-based models of
precedential constraint, a past decision binds later cases that are at least as strong for the side that won~\cite{horty2011rules,horty2012factor}, and CATO teaches arguing from the distinctions between cases~\cite{aleven2003using}. We study recorded labels that
can be wrong, under explicit rules.

\section{Design}
\label{sec:design}

\subsubsection{Audit procedure.}
(1) \emph{Select close calls}: requests that the model approves in 25\% to 75\% of decisions without precedents.
(2) \emph{Write close variants}: for each request, near-copies on each side whose facts call for approval or for
denial under the rules. (3) \emph{Swap one part at a time}: show the same variants with APPROVE labels, with DENY labels
and without labels, which gives $L$ and $F$. (4) \emph{Repeat with retrieval}: let a retriever choose the cases from a
store, and flip the stored labels to measure how far wrong outcomes move decisions.

\subsubsection{Decision task and models.}
Every decision is a single model call. The system message (the standing instruction before the conversation) names the organization, says ``Decide each request by
applying the written criteria to the verified facts'' and asks for JSON with a decision (APPROVE or DENY) and a
one-sentence reason. The target turn (the message that carries the request) gives the numbered criteria and a short fact sheet. Precedents are fact sheets in
the same format. Each is followed by the line \texttt{Decision: APPROVE}, the line \texttt{Decision: DENY} or no decision line. On the public policies the precedents arrive as the output of a lookup tool; on the fictional policies they are placed in the prompt before the target. The main model is DeepSeek
V4.1 Flash, with thinking (a written reasoning step before the answer) disabled and sampling temperature 1.0. The cross-model experiments use three Claude models (Haiku 5.5,
Sonnet 5.5, Opus 5.5) and three GPT models (GPT-6 Luna, GPT-6.1 Terra, GPT-6.1 Sol), the small, middle and large tier
of each family, at the medium setting for reasoning effort. A \emph{seed} is one repetition of a condition, with its own order of the
precedents and, where precedents are drawn from a larger pool, its own draw; retrieved cases keep their rank order. No API random seed is set.

\subsubsection{Materials.}
The authors, with an AI assistant, paraphrased nine public policies into 5--6 numbered criteria each (Steam, IRS
penalty relief, Airbnb, eBay, Best Buy, Apple warranty, AppleCare+, Allianz travel, Southwest) and wrote
\NCandidates{} candidate requests. The \NTargetsPublic{} candidates that DeepSeek approved in 25\% to 75\% of four
decisions without precedents became targets; in \BordCalls{} other decisions per target without precedents, \BordIn{} of the \NTargetsPublic{} targets stayed between 25\% and 75\%. For each target the authors wrote six \emph{original variants}, the target's fact sheet with
1--5 fields replaced, three whose facts call for approval and three for denial; we call this direction a variant's
\emph{reference label}. Original approve-side variants often depend on a different criterion than the target does, so each target also has six \emph{same-rule variants}, which change only the field or fields of the criterion that makes the
target a close call. Six fictional organizations have 4--5 criteria each; their \CThreeN{} targets have close
variants drafted by DeepSeek and kept only if DeepSeek decided them the intended way in two of two decisions. Table~\ref{tab:sets} compares the public and fictional materials.

\begin{table}[t]
\centering
\footnotesize
\caption{Materials for the public and the fictional policies.}
\label{tab:sets}
\begin{tabularx}{\linewidth}{@{}>{\raggedright\arraybackslash}p{3.3cm}>{\raggedright\arraybackslash}X>{\raggedright\arraybackslash}X@{}}
\toprule
& Public policies & Fictional policies\\
\midrule
Policies & 9 public policies, paraphrased & 6 fictional organizations\\
Targets & \NTargetsPublic{} & \CThreeN{}\\
Cases written by & the authors with an AI assistant & targets by the authors with an AI assistant or drafted by DeepSeek; close variants drafted by DeepSeek\\
Reference label (the direction the facts call for) & the authors' reading of the rules & DeepSeek's decision, the same in two of two calls\\
Precedents per target & 3 approve-side and 3 deny-side variants, fixed; a same-rule set of the same size & 3 per side, drawn for each seed from a pool of 3--6\\
\bottomrule
\end{tabularx}
\end{table}

\subsubsection{Conditions and measures.}
On the public policies each fact set is shown with all labels APPROVE, all labels DENY or no labels, besides the
decision without precedents; in the fictional experiment, with APPROVE and with DENY labels only. $L$ is approval with
APPROVE labels minus approval with DENY labels, averaged over the approve-side and the deny-side precedents. $F$ is approval with unlabeled
approve-side precedents minus approval with unlabeled deny-side precedents. Other experiments add one of two instructions about recorded decisions (the \emph{instruction} experiment); show the precedents as a user message, an assistant message or a tool result (\emph{delivery}); run with thinking enabled (\emph{thinking}); use an \emph{empty lookup} (the lookup tool runs but returns no cases, unlike the condition without precedents); show the close variants of another target of the same policy (a \emph{donor}); or let a retriever choose the cases (Section~\ref{sec:retrieval}). The fictional, delivery and thinking experiments use the six fictional policies; the others use the nine public policies.

\subsubsection{Human annotation and statistics.}
Three annotators, who took no part in writing the materials and were blind to the conditions and reference labels,
judged \AnnTargets{} targets, stratified by policy, and \AnnPrecedents{} original precedents, in random order, from the written policy alone; the final label of
each item was set by consensus. For all \AnnPrecAgree{} annotated precedents the consensus label equals the reference label. We call the consensus label of a target the \emph{human answer}; comparisons with it use the \AnnAccN{} targets with an APPROVE or DENY consensus. For
\AnnTargetNotClear{} of the \AnnTargets{} targets the policy does not directly decide the answer, so the human answer is a
consensus label on a close call. Intervals are 95\% bootstrap intervals that resample policies and seeds independently (10,000 draws); for comparisons with the human answer they resample targets instead (4,000 draws in Sections~\ref{sec:labels} and~\ref{sec:facts}). Every target and seed has equal weight in each estimate. An interval endpoint printed as 0.0 does not exclude zero. A difference is called \emph{detected} when its interval excludes zero. Target-level p-values come from a two-sided sign-flip test over targets and are Holm-adjusted~\cite{holm1979simple} within each experiment's families of tests; policy-level p-values come from an exact sign-flip test over the policies. A detected difference meets its criterion when its Holm-adjusted p is also below 0.05, with the predicted sign and a minimum size where these were fixed in advance: at least 8 points in the fictional and thinking experiments and 5 points for the delivery experiment's source differences. Measures, tests and criteria were fixed before the main model calls of
each experiment; comparisons with the human answer, the policy-level tests of the public-policy experiment, per-request counts and subgroup splits are exploratory.

\section{Results}
\label{sec:results}

\subsection{RQ1: Recorded labels pull decisions toward themselves in four of seven models}
\label{sec:labels}

\begin{figure}[!t]
\centering
\resizebox{\linewidth}{!}{\begin{tikzpicture}[font=\small]
\draw[gray!25] (-4.80,0.2) -- (-4.80,4.60);
\node[below, font=\footnotesize] at (-4.80,0.2) {$-$80};
\draw[gray!25] (-3.60,0.2) -- (-3.60,4.60);
\node[below, font=\footnotesize] at (-3.60,0.2) {$-$60};
\draw[gray!25] (-2.40,0.2) -- (-2.40,4.60);
\node[below, font=\footnotesize] at (-2.40,0.2) {$-$40};
\draw[gray!25] (-1.20,0.2) -- (-1.20,4.60);
\node[below, font=\footnotesize] at (-1.20,0.2) {$-$20};
\draw[gray!25] (0.00,0.2) -- (0.00,4.60);
\node[below, font=\footnotesize] at (0.00,0.2) {0};
\draw[gray!25] (1.20,0.2) -- (1.20,4.60);
\node[below, font=\footnotesize] at (1.20,0.2) {20};
\draw[gray!25] (2.40,0.2) -- (2.40,4.60);
\node[below, font=\footnotesize] at (2.40,0.2) {40};
\draw[gray!25] (3.60,0.2) -- (3.60,4.60);
\node[below, font=\footnotesize] at (3.60,0.2) {60};
\draw[gray!25] (4.80,0.2) -- (4.80,4.60);
\node[below, font=\footnotesize] at (4.80,0.2) {80};
\draw[gray] (0,0.2) -- (0,4.60);
\node[left, font=\footnotesize] at (-4.90,0.50) {AppleCare+};
\draw (0.812,0.58) -- (1.375,0.58);
\node[draw, circle, fill=black, inner sep=1.6pt] at (1.125,0.58) {};
\draw (-1.375,0.42) -- (-0.500,0.42);
\node[draw, rectangle, fill=white, inner sep=1.6pt] at (-1.000,0.42) {};
\node[left, font=\footnotesize] at (-4.90,1.00) {Apple warranty};
\draw (1.300,1.08) -- (2.000,1.08);
\node[draw, circle, fill=black, inner sep=1.6pt] at (1.600,1.08) {};
\draw (-3.000,0.92) -- (-0.200,0.92);
\node[draw, rectangle, fill=white, inner sep=1.6pt] at (-1.600,0.92) {};
\node[left, font=\footnotesize] at (-4.90,1.50) {Southwest};
\draw (0.800,1.58) -- (2.600,1.58);
\node[draw, circle, fill=black, inner sep=1.6pt] at (1.700,1.58) {};
\draw (-3.000,1.42) -- (-0.400,1.42);
\node[draw, rectangle, fill=white, inner sep=1.6pt] at (-1.600,1.42) {};
\node[left, font=\footnotesize] at (-4.90,2.00) {Allianz travel};
\draw (1.333,2.08) -- (2.167,2.08);
\node[draw, circle, fill=black, inner sep=1.6pt] at (1.750,2.08) {};
\draw (-1.833,1.92) -- (0.667,1.92);
\node[draw, rectangle, fill=white, inner sep=1.6pt] at (-0.500,1.92) {};
\node[left, font=\footnotesize] at (-4.90,2.50) {eBay};
\draw (1.417,2.58) -- (2.250,2.58);
\node[draw, circle, fill=black, inner sep=1.6pt] at (1.833,2.58) {};
\draw (-2.000,2.42) -- (-0.500,2.42);
\node[draw, rectangle, fill=white, inner sep=1.6pt] at (-1.167,2.42) {};
\node[left, font=\footnotesize] at (-4.90,3.00) {Best Buy};
\draw (2.062,3.08) -- (2.937,3.08);
\node[draw, circle, fill=black, inner sep=1.6pt] at (2.500,3.08) {};
\draw (-1.625,2.92) -- (0.625,2.92);
\node[draw, rectangle, fill=white, inner sep=1.6pt] at (-0.500,2.92) {};
\node[left, font=\footnotesize] at (-4.90,3.50) {Airbnb};
\draw (2.444,3.58) -- (2.889,3.58);
\node[draw, circle, fill=black, inner sep=1.6pt] at (2.667,3.58) {};
\draw (-4.222,3.42) -- (-3.222,3.42);
\node[draw, rectangle, fill=white, inner sep=1.6pt] at (-3.667,3.42) {};
\node[left, font=\footnotesize] at (-4.90,4.00) {Steam};
\draw (2.857,4.08) -- (3.357,4.08);
\node[draw, circle, fill=black, inner sep=1.6pt] at (3.143,4.08) {};
\draw (-1.857,3.92) -- (-0.286,3.92);
\node[draw, rectangle, fill=white, inner sep=1.6pt] at (-1.000,3.92) {};
\node[left, font=\footnotesize] at (-4.90,4.50) {IRS penalty relief};
\draw (3.091,4.58) -- (4.091,4.58);
\node[draw, circle, fill=black, inner sep=1.6pt] at (3.591,4.58) {};
\draw (-1.818,4.42) -- (-0.909,4.42);
\node[draw, rectangle, fill=white, inner sep=1.6pt] at (-1.364,4.42) {};
\node[below] at (0,-0.25) {Effect on target approval rate (points)};
\node[draw, circle, fill=black, inner sep=1.6pt] at (-3.0,4.95) {};
\node[right, font=\footnotesize] at (-2.9,4.95) {label effect $L$};
\node[draw, rectangle, fill=white, inner sep=1.6pt] at (0.6,4.95) {};
\node[right, font=\footnotesize] at (0.7,4.95) {facts contrast $F$};
\end{tikzpicture}}
\caption{For DeepSeek, in every public policy the label effect is positive and the point estimate of the facts contrast negative. Per-policy
$L$ (filled circles) and $F$ (open squares), DeepSeek V4.1 Flash, original variants. Bars are 95\% bootstrap intervals over seeds within
the policy. Each policy has \TargetsPerPolicyMin{}--\TargetsPerPolicyMax{} targets and \RFSeeds{} seeds.}
\label{fig:policy}
\end{figure}

Changing only the three labels from DENY to APPROVE raises DeepSeek's approval by \RFLabelPt{} points on the nine
public policies (Table~\ref{tab:main}). The effect is positive in each of the nine policies (Figure~\ref{fig:policy}; exact sign-flip test over policies, p = \RFLabelPolP{}).
It is \SRLabelPt{} points with the same-rule variants and \CThreeDeepPt{} points on the fictional policies, where Claude Haiku 5.5 gives \CThreeHaiku{}. In the \BordCalls{} other decisions per target without precedents (Section~\ref{sec:design}), \BordOut{} targets fell outside 25\% to 75\% approval; the label effect was \BordLOut{} points on those and \BordLIn{} on the other \BordIn{}. In the delivery experiment it stays
large whether the precedents arrive as a user message, an assistant message or a tool result (\ETwoLabelToolEst{} to
\ETwoLabelAsstEst{} points), and in the thinking experiment it is \ThinkOn{} points with thinking enabled against \ThinkOff{} without; thinking also lowered approval without precedents.

An instruction lowers the label effect without removing it. In the instruction experiment's own run without an added sentence, the label effect was \ILBase{} points. Adding ``Past decisions in the database may have been
recorded wrongly; the written criteria decide.'' to the system message lowered it to \ILWarn{} (a change of \IdLWarn{}); adding
``Do not rely on the recorded decisions of past cases; decide only from the written criteria and the verified facts of
the current request.'' lowered it to \ILIgnore{} (a change of \IdLIgnore{}).

\begin{table}[!t]
\centering
\footnotesize
\caption{APPROVE labels raise approval; with the original variants, approve-side facts give lower approval than
deny-side facts. Top: approval rate (\%) of the target by the labels and facts of its three original variants; DeepSeek V4.1
Flash, public policies, \RFCellN{} decisions per cell; without precedents \RFRateNone{}\%. Bottom: $L$ and $F$ in
points with 95\% intervals. $L$ is the APPROVE row minus the DENY row, averaged over the two columns; $F$ is the `none' row, approve-side minus deny-side column. $F$ is not defined in the fictional experiment.}
\label{tab:main}
\begin{tabular}{@{}lcc@{}}
\toprule
Labels on the three precedents & Approve-side facts & Deny-side facts\\
\midrule
APPROVE & \RFRateApproveStrong{} & \RFRateApproveWeak{}\\
none & \RFRateOmittedStrong{} & \RFRateOmittedWeak{}\\
DENY & \RFRateRejectStrong{} & \RFRateRejectWeak{}\\
\bottomrule
\end{tabular}

\vspace{6pt}
\begin{tabular}{@{}lc@{}}
\toprule
& DeepSeek\\
\midrule
$L$, original variants & \RFLabel{}\\
$F$, original variants & \RFFacts{}\\
$L$, same-rule variants & \SRLabel{}\\
$F$, same-rule variants & \SRFacts{}\\
$L$, fictional policies & \CThreeDeep{}\\
\bottomrule
\end{tabular}
\end{table}

\begin{table}[t]
\centering
\footnotesize
\caption{The label effect differs between models; the facts contrast is negative in all of them. $L$ and $F$ with the
original and the same-rule variants, \NTargetsPublic{} public-policy targets, \ModSeeds{} seeds (\ModDecisions{}
decisions per model). Points with 95\% intervals; $^{*}$: interval excludes zero and two-sided Holm p $<$ 0.05; an endpoint printed as 0.0 does not exclude zero.}
\label{tab:models}
\setlength{\tabcolsep}{4pt}
\resizebox{\linewidth}{!}{%
\begin{tabular}{@{}lcccc@{}}
\toprule
& $L$, original & $L$, same-rule & $F$, original & $F$, same-rule\\
\midrule
Claude Haiku 5.5 & +28.2 [20.1, 35.6]$^{*}$ & +31.3 [22.2, 39.0]$^{*}$ & $-$22.1 [$-$32.3, $-$12.6]$^{*}$ & $-$13.3 [$-$24.0, $-$2.5]$^{*}$\\
Claude Sonnet 5.5 & +11.0 [2.2, 21.5]$^{*}$ & +23.1 [13.6, 30.2]$^{*}$ & $-$27.2 [$-$38.9, $-$15.9]$^{*}$ & $-$31.3 [$-$42.1, $-$18.8]$^{*}$\\
Claude Opus 5.5 & $-$4.4 [$-$9.0, 0.0] & $-$3.6 [$-$9.3, 3.7] & $-$19.0 [$-$28.1, $-$9.6]$^{*}$ & $-$40.5 [$-$50.7, $-$29.5]$^{*}$\\
\midrule
GPT-6 Luna & +23.3 [9.4, 34.9]$^{*}$ & +36.4 [26.3, 44.8]$^{*}$ & $-$19.5 [$-$32.8, $-$7.2]$^{*}$ & $-$30.3 [$-$46.9, $-$11.3]$^{*}$\\
GPT-6.1 Terra & +3.8 [$-$3.1, 10.8] & +2.3 [$-$3.0, 8.5] & $-$12.8 [$-$24.2, $-$4.2]$^{*}$ & $-$25.1 [$-$40.6, $-$10.2]$^{*}$\\
GPT-6.1 Sol & $-$3.1 [$-$7.7, 1.9] & $-$7.9 [$-$12.1, $-$4.3]$^{*}$ & $-$27.2 [$-$44.8, $-$11.6]$^{*}$ & $-$41.0 [$-$55.3, $-$24.8]$^{*}$\\
\bottomrule

\end{tabular}}
\end{table}

The label effect differs between models (Table~\ref{tab:models}). Claude Haiku 5.5, Claude Sonnet 5.5 and GPT-6 Luna
show it with both the original and the same-rule variants. For Claude Opus 5.5 and GPT-6.1 Terra neither label effect was detected, and for
GPT-6.1 Sol the label effect with the same-rule variants was negative. Averaged over the original and same-rule variants, the label effect of
Opus minus that of Haiku was \BtwOpusHaikuLPt{} points and that of Sol minus Luna \BtwSolLunaSixLPt{} points. A label
effect near zero is an average over requests in which opposite shifts can cancel. With the same-rule variants, APPROVE
labels on balance raised Opus's approval for \UpOpusSr{} requests and lowered it for \DownOpusSr{}.

On the annotated targets, relative to no precedents, DeepSeek's agreement with the human answer changed by \AnnTowardDiff{} points when the three labels matched the human answer and
by \AnnAwayDiff{} points when they contradicted it.

\subsection{RQ2: Unlabeled facts move the decision in all six Claude and GPT models}
\label{sec:facts}

With the original variants and no labels, approve-side precedents give lower approval than deny-side ones: $F$ is
\RFFactsPt{} points for DeepSeek, negative in all nine policies (p = \RFFactsPolP{}). With the same-rule variants it was
\SRFacts{} points and not detected. Relative to an empty lookup, unlabeled approve-side precedents changed DeepSeek's
approval by \XAOrigEmpty{} points with the original and \XASrEmpty{} with the same-rule variants. In the six Claude and
GPT models the facts contrast was negative with both sets of variants in all \ModFCells{} cases (Table~\ref{tab:models}),
including Claude Opus 5.5, GPT-6.1 Terra and GPT-6.1 Sol, for which no positive label effect was detected. Relative to the empty lookup, unlabeled approve-side precedents gave decreases that meet their criteria in all six models with the same-rule variants, and unlabeled deny-side precedents gave such increases with at least one set of variants in all models except GPT-6.1 Terra. GPT-6.1 Sol, which matched the human
answer in \ModSolHumNone{}\% of decisions without precedents, agreed less often than without precedents in each of the four unlabeled conditions (approve-side or deny-side precedents, original or same-rule variants).

Two explanations fit these results. In \emph{discriminative inference}, the model reads from approve-side precedents what the target lacks, whether or not the rules make it decisive. In a \emph{contrast bias}, approve-side precedents raise the bar for approval. On the annotated targets, DeepSeek's facts contrast with the original variants was \AnnFactsHumApp{} points where the human consensus was APPROVE and \AnnFactsHumDeny{} where it was DENY, and it was not detected to differ between the two, so these subgroup results do not distinguish the explanations. Under either, a system cannot assume that a precedent's facts move the decision in the direction of its label.

\subsection{RQ3: Own close variants carry the label effect}
\label{sec:boundary}

The own-and-donor experiment showed each public-policy target its own close variants and, in separate conditions, those
of a \emph{donor}, another target of the same policy assigned at random (DeepSeek, \ODSeeds{} seeds). With the donor's
variants the label effect was \ODLDonor{} points and not detected, against \ODLOwn{} with the target's own; the
difference was \ODDL{} (Table~\ref{tab:owndonor}). The facts contrast remained with donor precedents (\ODFDonor{} points, against \ODFOwn{}).

\begin{table}[t]
\centering
\footnotesize
\caption{Another request's precedents lose most of the label effect and keep a facts contrast.
Own-and-donor experiment, DeepSeek V4.1 Flash, \NTargetsPublic{} public-policy targets $\times$ \ODSeeds{} seeds. Points with
95\% intervals; below each row, the Holm-adjusted p within its family and, in parentheses, the exact policy-level p. An effect is detected only when its interval excludes zero, even when its p is below 0.05.\vspace{3pt}}
\label{tab:owndonor}
\resizebox{\linewidth}{!}{\begin{tabular}{@{}lccc@{}}
\toprule
& Own close variants & Donor's close variants & Own minus donor\\
\midrule
Label effect $L$ & \ODLOwn{} & \ODLDonor{} & \ODDL{}\\
\quad Holm p (policy-level p) & \ODLOwnHolm{} (\ODLOwnPolP{}) & \ODLDonorHolm{} (\ODLDonorPolP{}) & \ODDLHolm{} (\ODDLPolP{})\\
Facts contrast $F$ & \ODFOwn{} & \ODFDonor{} & \ODDF{}\\
\quad Holm p (policy-level p) & \ODFOwnHolm{} (\ODFOwnPolP{}) & \ODFDonorHolm{} (\ODFDonorPolP{}) & \ODDFHolm{} (\ODDFPolP{})\\
\bottomrule
\end{tabular}}
\end{table}

\subsection{RQ4: With retrieved cases, wrong stored labels move DeepSeek's decisions}
\label{sec:retrieval}

A store holds all \RetStore{} public-policy variants (the original and same-rule variants of the \NTargetsPublic{} targets) with their reference labels. For each target, keyword search
(BM25~\cite{robertson2009bm25}) over the fact sheets returns the three highest-ranked cases of the target's policy; for
\RetOwnAll{} of the \NTargetsPublic{} targets all three are the target's own variants. A second store leaves out the
target's own variants. The retrieved cases are shown with their reference labels, with every label flipped, without
labels, or with all labels APPROVE or all DENY.

\begin{table}[t]
\centering
\footnotesize
\caption{With retrieved cases, labels move DeepSeek's decisions and wrong labels propagate. \NTargetsPublic{}
targets; DeepSeek \RetDSSeeds{} seeds, Opus and GPT-6.1 Sol \ModSeeds{} seeds. $L$: approval with all retrieved labels
APPROVE minus all DENY. $W$: change in approval when every stored label is flipped, signed positive when it moves toward the majority of the three flipped labels.}
\label{tab:retrieval}
\setlength{\tabcolsep}{4pt}
\resizebox{\linewidth}{!}{%
\begin{tabular}{@{}lcc@{}}
\toprule
& $L$ & $W$\\
\midrule
DeepSeek V4.1 Flash, full store & +71.0 [57.5, 82.6] & +24.9 [11.9, 37.2]\\
DeepSeek V4.1 Flash, store without the target's own variants & +25.1 [16.1, 34.5] & +15.6 [5.9, 25.9]\\
\midrule
Claude Opus 5.5, full store & +11.3 [0.0, 19.0] & +1.5 [$-$7.6, 10.4]\\
GPT-6.1 Sol, full store & $-$1.5 [$-$11.7, 8.1] & $-$1.5 [$-$8.9, 4.8]\\
\bottomrule

\end{tabular}}
\end{table}

DeepSeek's label effect was \RetDSLFullPt{} points with the full store and \RetDSLLooPt{} without the target's own
variants (Table~\ref{tab:retrieval}). Flipping every stored label shifted DeepSeek's approval by \RetDSWFullPt{} points
toward the flipped labels ($W$ in Table~\ref{tab:retrieval}), and by \RetDSWLooPt{} without the target's own variants. For Claude Opus 5.5 and GPT-6.1 Sol neither a label effect nor a shift toward flipped
labels was detected. A shift toward flipped labels is not the same as a wrong decision on the target. On the annotated targets, with the full store and relative to no precedents, retrieved cases with their reference labels changed the agreement of Opus and Sol with the human answer by \RetHumOpusRefMinusNone{} and \RetHumSolRefMinusNone{} points; relative to those reference labels, flipping the labels changed it by \RetHumOpusFlipMinusRef{} and \RetHumSolFlipMinusRef{}, and for DeepSeek by \RetHumDSFlipMinusRef{}.

\section{Implications and limitations}
\label{sec:implications}

Four checks follow for a system that shows a language model past decisions next to written rules.
\begin{enumerate}
\item Stored labels can decide close calls. In four of the seven models the labels of close variants moved approval
toward themselves, and in the constructed case store wrong stored labels propagated to DeepSeek's decisions.
\item An instruction does not neutralize stored labels. Neither of the two instructions tested removed DeepSeek's label effect (Section~\ref{sec:labels}).
\item Unlabeled cases are not neutral. Their facts moved the decision in all six Claude and GPT models.
\item Both effects need testing again when the model changes, because their sizes differed between models
of one family.
\end{enumerate}

The results cover the models, prompts and settings tested, with a single-call decision on a short fact sheet. The
public-policy requests, variants and reference labels were written by the authors with an AI assistant, under
paraphrased policies, and targets were selected as close calls for DeepSeek, so the estimates for other models are
conditional on that selection. With nine policies, several effects that meet their criteria have policy-level p-values above 0.05. The small, middle and large models of a family differ in more than capability. Each cross-model run has three
seeds. The retrieval store holds variants written for this study; retrieval from records of real requests remains to be
studied.

\section{Conclusion}
\label{sec:conclusion}

On requests under written rules, the recorded labels and the facts of similar precedents act differently, and the
sizes of their effects differ between models. Four of seven models follow the labels of close variants, DeepSeek also when instructed
not to rely on them, while for Claude Opus 5.5 and GPT-6.1 Sol no positive label effect was detected and Sol moved away from the labels with the same-rule variants. Unlabeled
precedents whose facts call for approval gave lower approval than those whose facts call for denial in all six Claude
and GPT models.

\begin{credits}
\subsubsection{\discintname}
The authors have no competing interests to declare that are relevant to the content of this article. This research
did not receive any specific grant from funding agencies in the public, commercial, or not-for-profit sectors.
\end{credits}

\bibliographystyle{splncs04}
\bibliography{references}

\end{document}